# The importance of space and time in neuromorphic cognitive agents

Giacomo Indiveri, *Senior Member, IEEE,* Yulia Sandamirskaya, *Member, IEEE*

## Abstract

Artificial neural networks and computational neuroscience models have made tremendous progress, allowing computers to achieve impressive results in artificial intelligence (AI) applications, such as image recognition, natural language processing, or autonomous driving. Despite this remarkable progress, biological neural systems consume orders of magnitude less energy than today's artificial neural networks and are much more agile and adaptive. This efficiency and adaptivity gap is partially explained by the computing substrate of biological neural processing systems that is fundamentally different from the way today's computers are built. Biological systems use in-memory computing elements operating in a massively parallel way rather than time-multiplexed computing units that are reused in a sequential fashion. Moreover, activity of biological neurons follows continuous-time dynamics in real, physical time, instead of operating on discrete temporal cycles abstracted away from real-time. Here, we present neuromorphic processing devices that emulate the biological style of processing by using parallel instances of mixed-signal analog/digital circuits that operate in real time. We argue that this approach brings significant advantages in efficiency of computation. We show examples of embodied neuromorphic agents that use such devices to interact with the environment and exhibit autonomous learning.

## I. INTRODUCTION

Tremendous progress in both machine learning and computational neuroscience is leading to the development of neural processing algorithms that have far reaching impact on our daily lives [1]. These neural network algorithms can perform remarkable learning and pattern recognition tasks that in some cases even outperform humans. However, they are typically run on conventional computing systems based on the von Neumann architecture and are commonly

G. Indiveri and Y. Sandamirskaya are with the Institute of Neuroinformatics, University of Zurich and ETH Zurich, Zurich, Switzerland.

implemented using large and power-hungry platforms, sometimes distributed across multiple machines in server farms. The power required to run these algorithms and achieve impressive results is orders of magnitude larger than the one used by biological nervous systems and is not sustainable for the future needs of ubiquitous computing at scale. The reasons for the large gap in energy costs between artificial and natural neural processing systems are still being investigated, however it is clear that one fundamental difference lies in the way the elementary computing processes are organized: conventional computers use Boolean logic, bit-precise digital representations, time-multiplexed and clocked operations; nervous systems, on the other hand, carry out robust and reliable computation using analog components that are inherently noisy and operate in continuous time and activation domains. These components typically communicate among each other with all-or-none discrete events (spikes), thus using a combination of analog computation and digital communication. Moreover, they form distributed, event-driven, and massively parallel systems; and they feature considerable amount of adaptation, self-organization, and learning with dynamics that operate on a multitude of time-scales.

One promising approach for bridging this gap is to develop a new generation of ultra-low power and massively parallel computing technologies, optimally suited to implementing neural network architectures that use the same spatial and temporal principles of computation used by the brain. This "neuromorphic engineering" approach [2] is the same one that was originally proposed in the early '90s [3] but which is starting to show its full potential only now. Neuromorphic circuits are typically designed using analog, digital, or mixed-mode analog/digital Complementary Metal-Oxide-Semiconductor (CMOS) transistors, and are well suited for exploiting the features of emerging memory technologies and new memristive materials [4–6]. Similar to the biological systems they model, neuromorphic systems process information using energy-efficient asynchronous, event-driven methods [7]; they are often adaptive, fault-tolerant, and can be flexibly configured to display complex behaviors by combining multiple instances of simpler elements.

Remarkable neuromorphic computing platforms have been developed in the past for modeling cortical circuits, solving pattern recognition problems, and implementing machine learning tasks [8–19] or for accelerating the simulation of computational neuroscience models [15, 17]. In parallel, impressive full-custom dedicated accelerators have been proposed for implementing convolutional and deep network algorithms following the more conventional digital design flow [20, 21]. While these systems significantly reduce the power consumption in a wide variety of neural network applications, compared to conventional computing approaches, they still fall short of reproducing the power, size, and adaptivity of biological neural processing systems.

Indeed, developing low power, compact, and autonomous electronic "agents" that can interact with the environment to extract relevant information from the signals they sense in real-time, adapt to the changes and uncertain conditions present in both the external inputs and internal states, and that can learn to produce context dependent behaviors for carrying out goal-directed procedural tasks, is still an open challenge.

To address this challenge and find an optimal computational substrate that minimizes size and power consumption for systems that generate behavior in real world, we argue, the computing principles that underlie autonomous adaptive behavior need to be matched by the computing hardware in which computation is realized. In particular, in order to fully benefit from the emulation of biological neural processing systems, it is important to preserve two of their fundamental characteristics: the explicit representation of time and the explicit use of space. In the following, we will first present the design principles for building neuromorphic electronic systems that make use of these representations, and then show how such explicit representation of time and space matches a computing framework of dynamic neural fields that embodies principles of autonomous behavior generation and learning [22, 23]. Finally, we demonstrate successful realizations of autonomous neural-dynamic architectures in neuromorphic hardware, emphasizing the role of explicit time and space representation in hardware for efficiency of the autonomous neuronal controllers.

## II. Design principles for building biologically plausible neuromorphic processors

### A. Physical space in hardware neural processing systems

In an effort to minimize silicon area consumption, digital neuromorphic processors typically use time-multiplexing techniques to share circuits which simulate neural dynamics for modeling multiple neurons [16, 18, 19]. This requires that the shared circuits continuously transfer their state variables to and from an external memory block at each update cycle (therefore burning extra power for the data transfer). The faster the transfer and cycle rate, the larger number of neurons can be simulated per time unit. In addition, if these circuits need to model processes that evolve continuously over natural time, such as the leak term of a leaky integrate and fire (I&F) neuron model, then it is necessary to include a clock to update the related state variables periodically and manage the passing of time (thus adding extra overhead and power consumption).

Unlike digital simulators of neural networks, analog neuromorphic circuits use the physics of silicon to directly emulate neural and synaptic dynamics [2]. In this case the state variables evolve naturally over time and "time represents itself" [3], bypassing the need to have clocks and extra circuits to manage the representation of time. Furthermore, since the state variable memory is held in the synapse and neuron capacitors there is no need to transfer data to extra memory blocks, dramatically saving energy that would otherwise be required to transfer the neuron state-variables back and forth from memory. Examples of neuromorphic architectures that follow this mixed-signal approach include the Italian ISS recurrent network with plasticity and long-term memory chip [8], the ISS "final learning attractor neural network" (FLANN) chip [9], the Georgiatech Learning-Enabled Neuron Array IC [10], the UCSD integrate-and-fire array transceiver (IFAT) architecture [11], the US Stanford Neurogrid system [12], the Zurich INI dynamic neuromorphic asynchronous processor chip [13], and the Zurich INI recurrent on-line learning ROLLS neuromorphic processor [14].

In these devices the analog synapse and neuron circuits have no active components [2]. The circuits are driven directly by the input "streaming" data. Their synapses receive input spikes and their neurons produce output spikes, at the rate of the incoming data. So if they are not processing data, there is no energy dissipated per synaptic operation (SOP) and no dynamic power consumption. Therefore, this approach is particularly attractive in the case of applications in which the signals have sparse activity in space and time. Under these conditions most neurons would be silent at any one time, thus bringing the system power consumption to the minimum.

A quantitative comparison of the specification of these devices is presented in Table I. The spike-based learning algorithms that some of these devices implement are either based on the basic version of Spike-Timing Dependent Plasticity (STDP) [24], or on more elaborate Spike-Timing and Rate Dependent Plasticity (STRDP) [25].

While this approach is very power efficient, it requires to instantiate and place a distinct physical neuromorphic circuit per emulated synapse or neuron, therefore needing an amount of physical area that scales with the number of neurons in the network. This was a serious limiting factor with older VLSI fabrication processes. However, technology scaling pushed by Moore's law and the emergence of novel nano-scale memory devices technologies that can be used to implement both synaptic and neural dynamics [5, 6] brings renewed interest to this approach and makes it very competitive compared to the classical one of resorting to more compact digital designs based on shared time-multiplexed circuits, as was done, e.g., for the IBM TrueNorth [16] and Intel Loihi [19] neuromorphic processors.

TABLE I: Quantitative comparison of mixed-signal neuromorphic processor specifications.

|  | [8] | [9] | [10] | [11] | [12] | [13] | [14] |
| --- | --- | --- | --- | --- | --- | --- | --- |
| **Technology** | 0.6 $\mu$m | 0.6 $\mu$m | 0.35 $\mu$m | 90 nm | 0.18 $\mu$m | 0.18 $\mu$m | 0.18 $\mu$m |
| **Supply voltage** | 3.3 V | 3.3 V | 2.4 V | 1.2 V | 3 V | 1.8 V | 1.8 V |
| **Core area** | 10 mm$^2$ | 68.9 mm$^2$ | 25 mm$^2$ | 139 $\mu$m$^2$ | 170 mm$^2$ | 7.5 mm$^2$ | 51.4 mm$^2$ |
| **Neurons/core** | 21 | 128 | 100 | 2000 | 65636 | 256 | 256 |
| **Synapses/core** | 129 | 16384 | 30000 | N/A | 4096 | 16000 | 128000 |
| **Fan-in/Fan-out** | 21/21 | 128/128 | 100/100 | N/A | N/A | 64/4000 | 256/256 |
| **Synaptic weight** | Capacitor | Capacitor | >10-bit | 8-bit | 13-bit shared | 1+1 bit | Capacitor |
| **On-line learning** | STRDP | STRDP | STDP | No | No | No | STRDP |
| **Energy per SOP** | N/A | N/A | 10 pJ | 22 pJ | 31.2 pJ | 17 pJ | 77 fJ |

*B. Natural time in hardware neural processing systems*

For the approach based on parallel instances of mixed analog/digital circuits as described above, the most power efficient way of processing time varying signals is to use circuit time constants that are well matched to those of the dynamics of the signals that need to be processed. For example, real-world "natural" events and signals such as speech, bio-signals measured from the body, human gestures, or motor and sensory signals measured from roving robots, would require the synapse and neural circuits to have time constants in the range of 5 ms to 500 ms. It is important to realize that although the speed of the individual synapse or neuron computing elements in such architectures can be set to be relatively slow (e.g., compared to digital circuits), the response time of the overall system can be extremely fast. This is due to the fact that the many parallel processing nodes in the system will be affected by device mismatch and have different initial conditions; so upon the arrival of an external input there will be many neurons very close to their firing threshold, and they will produce an output with a latency that is much shorter than their natural integration time-constant.

The relatively long time constants required by this approach are not easy to realize using analog CMOS technology. Standard analog circuit design techniques either lead to bulky and silicon-area expensive solutions or fail to meet this condition, resorting to modeling neural dynamics at accelerated time-scales [26]. One elegant solution to this problem is to combine the use of subthreshold circuit design techniques [3] with current-mode design one [27]. A very compact subthreshold log-domain circuit that can reproduce biologically plausible synaptic and neural temporal dynamics and that has been used in a wide variety of neuromorphic applications [25, 28–30] is the Differential Pair Integrator (DPI) circuit [31], depicted in Fig. 1. The analytic transfer function of this circuit can be derived following a translinear-loop log-domain analysis [2, 27]: From Fig. 1b it follows that

$$\tau \left(1 + \frac{I_g}{I_{out}}\right) \frac{d}{dt} I_{out} + I_{out} = \frac{I_g I_{in}}{I_\tau} - I_g, \tag{1}$$

where $I_{out}$ represents the synaptic or neuron dynamic variable, $I_{in}$ the input signal, $I_\tau$ a user defined leakage current, and $I_g$ a global gain term useful e.g. for modeling spike-based learning, intrinsic plasticity, or synaptic scaling. This is a non-linear differential equation. However, under the reasonable assumptions of non-negligible input currents, such that $I_{in} \gg I_\tau$, and observing that for such condition the output current $I_{out}$ will eventually grow to be $I_{out} \gg I_g$, then this equation simplifies to:

$$\tau \frac{d}{dt} I_{out} + I_{out} = \frac{I_g}{I_\tau} I_{in}. \tag{2}$$

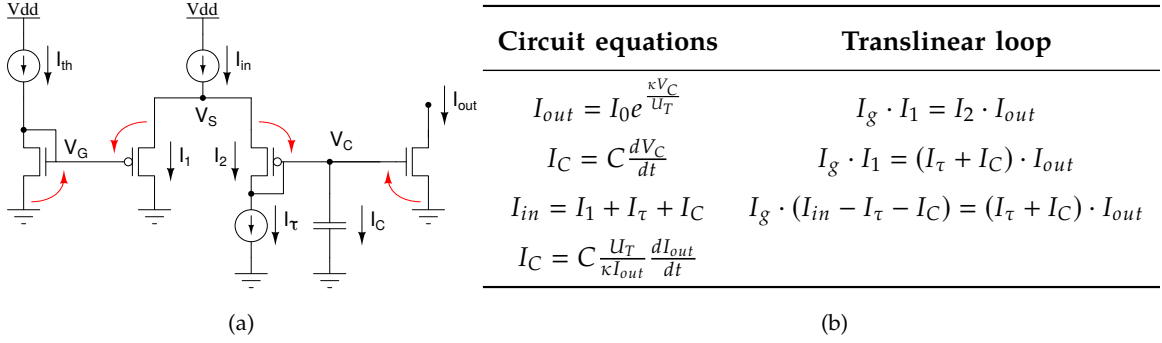

(a) (b)

Fig. 1: (a) Differential Pair Integrator circuit diagram; red arrows show the translinear loop considered for the log-domain analysis. (b) Circuit and translinear loop equations used to derive the circuit transfer function. The term $I_0$ in the equation represents the transistor dark current, $U_T$ the thermal voltage and $\kappa$ the subthreshold slope factor [3].

The circuit time constant is defined ad $\tau \triangleq CU_T/\kappa I_\tau$. Long time constants are achieved by using a combination of large capacitors and small currents. Even though it is possible to implement such long time constants without sacrificing large amounts of silicon real-estate area, e.g., using high-k materials, memristive devices, or active circuits that minimize leakage currents [30], there is a limit to the maximum time constants that can be implemented at the level of the single neural or synaptic components, and to the temporal scales that they can deal with for processing slow changing signals. This is a problem that also biological neural processing systems face, and which can be solved by using network dynamics to model attractors and long-term memory phenomena.

*C. Extending neuronal dynamics to behavioral time scales using attractor networks*

Biological nervous systems are capable of controlling behavior and processing sensory inputs in an adaptive and robust manner over a multitude of time scales that extend well beyond those of the single synapse and neuron time constants [32]. Indeed in such systems there are a multitude of neural circuits and mechanisms that underlie the ability to process and generate temporal patterns over behavioral time-scales [33]. A common circuit motive found throughout multiple parts of the cortex that is believed to sub-serve these functions is the "winner-take-all" (WTA) network [34, 35], which is a specific type of attractor network [36]. Theoretical studies have shown that such networks provide elementary units of computation that can stabilize and de-noise the neuronal dynamics [34, 37, 38]. These theoretical considerations have also been validated in neuromorphic hardware systems to generate robust behavior in closed sensorimotor loops [39]. However, to extend these models and hardware neural processing circuits to more complex systems, such as autonomous agents that can make decisions and generate goal-directed behavior, it is necessary to develop higher-level control strategies and theoretical frameworks compatible with mixed signal neuromorphic hardware and endowing neuromorphic architectures with compositionality and modularity. The core challenge is to design neuronal networks for neuromorphic hardware that can create and stabilize a neuronal state that can drive movements of an autonomous agent that unfold at arbitrary time-scales.

Recurrent networks of neural populations can indeed create sustained activation to keep a neuronal state active for macroscopic, behavioral time intervals. Such sustained neuronal states do not have to be static, but need to create an invariant projection in the task-related subspace [40], creating an attractor in this subspace that corresponds to the desired value (i.e., a working memory state [41]). These attractor-based representations can sustain a variable duration of actions and perceptual states and help to bridge the neuronal and behavioral time scales in a robust way [23, 41]. For example, a recurrent neural network dynamics that can be used for this purpose is described by the following equation [22]:

$$\tau \dot{u}(x,t) = -u(x,t) + h + I(x,t) + \int f(u(x',t))w(|x-x'|)dx'. \qquad (3)$$

Here, $u(x,t)$ is a neuronal activation function defined over a perceptual or motor space $x$ that is encoded by a neuronal population. Such dimension, $x$, could represent a continuous sensory feature such as color or orientation of a visual stimulus, or a motor variable such as the hand position or orientation in space. The term $h$ in Eq. (3) is a negative resting level of the neuronal population; the term $I(x, t)$ is an external input, and $f(\cdot)$ is a sigmoid non-linearity that smoothly filters population activity. The last, integral term in the equation expresses the lateral connectivity in the neuronal population with activity described by the activation function $u(x, t)$. In particular, the integral is a convolution shaped by the interaction kernel $w(|x - x'|)$ that only depends on the distance between two positions. The interaction kernel has a "Mexican hat" form of the type:

$$w(|x - x'|) = A_{exc} e^{\frac{(x-x')^2}{2\sigma_{exc}^2}} - A_{inh} e^{\frac{(x-x')^2}{2\sigma_{inh}^2}}, \quad (4)$$

where $A_{exc/inh}$ and $\sigma_{exc/inh}$ are the amplitude and spread of the excitatory and inhibitory parts of the kernel. Such "Mexican hat" pattern of lateral connections creates a soft WTA dynamics in the neuronal population: the neurons that get activated first, if they are supported by neighbouring (in the behavioral space) neurons, stay activated and inhibit other neurons in the population. This formalization of the dynamics of recurrent neural populations is known as a Dynamic Neural Field (DNF) [42, 43]. Over decades, DNF theory was developed into a framework for neuronal basis of cognition [22] that has been recently applied to the control of cognitive robotic agents [23, 44].

The Dynamic Neural Field dynamics described by Eq. (3) can be implemented as a WTA network in neuromorphic hardware. Figure 2a shows such a scheme of a WTA / DNF implementation with spiking neurons. Here, red circles designate neurons, a larger number of which form an excitatory pool that represents the behavioral variable; the smaller pool of neurons forms an inhibitory group that realises the inhibitory part of interaction kernel in Eq. (3). Red lines are excitatory and blue lines are inhibitory connections (shown for one of the inhibitory neurons each).

The stable attractors created by such WTA dynamics are critical to enable behavior learning in a closed sensory-motor loop in which the sensory input changes continually as the agent generates action. In order to learn a mapping between a sensory state and its consequences, or a precondition and an action, the sensory state before the action needs to be stored in a neuronal representation. This can be achieved by creating a reverberating activation in a neuronal population that can be sustained for the duration of the action and can be used to update the sensorimotor mapping when a rewarding or punishing signal is obtained [44, 45].

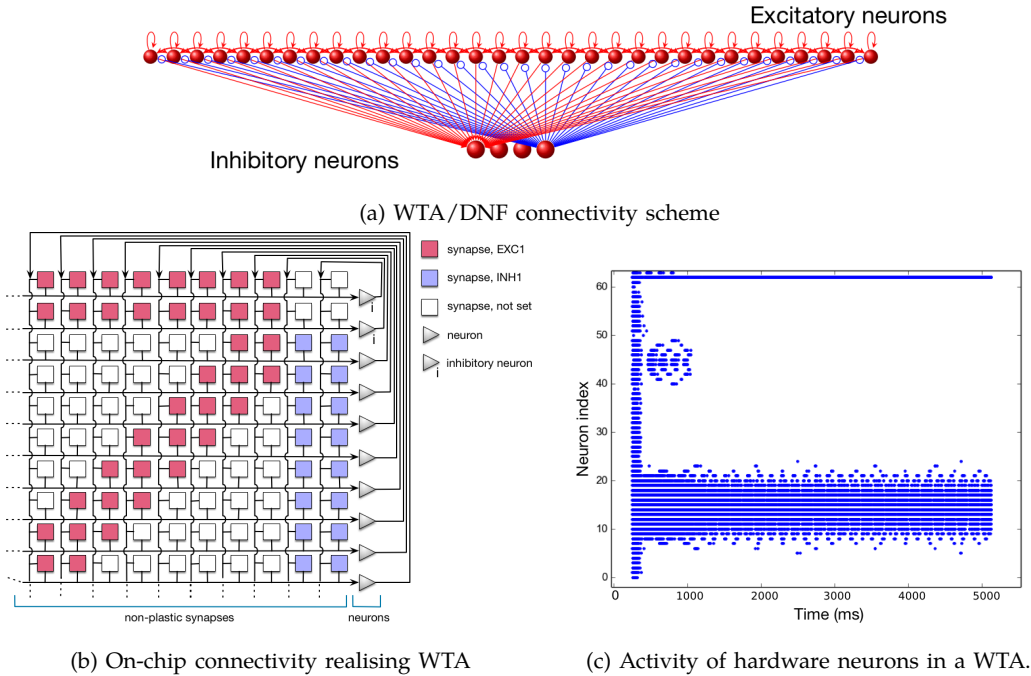

Fig. 2: (a): Neuronal connectivity realising a WTA / Dynamic neural fields (red are excitatory connections, blue are inhibitory connections; only one set of excitatory - to - inhibitory and inhibitory - to - excitatory connections are shown. (b): The same connectivity realized in non-plastic synapses on a mixed-signal neuromorphic processor. (c): Illustration of noise-reducing properties of the DNF dynamics – raster plot of spiking activity in a WTA architecture implemented on a neuromorphic processor [14].

### D. Implementing on-chip learning in neuromorphic processors

Learning is probably the most important aspect of neural processing systems: it allows them to be trained to form memories, create associations, perform pattern classification or pattern recognition tasks; most importantly, it endows autonomous agents with plasticity features which enable them to adapt to changes in the statistics of the input signals, or changes in the properties of their actuators over long time scales.

In artificial neural processing systems learning typically involves the update of synaptic weight values. This can require significant amounts of extra resources in time-multiplexed systems, typically in the form of state memory, memory-transfer bandwidth, and power, especially if the storage is done using memory banks that are not on the processor die (e.g., DRAM in large-scale systems). On the other hand, the overhead needed to implement learning in mixed-signal neuromorphic architectures that place multiple instances of synaptic circuits per emulated synapse is very small. For example, Fig. 3, depicting the chip micrograph of the ROLLS device [14], shows how the synapses that comprise learning algorithms (denoted as LTP synapses) occupy approximately the same area of the non-plastic synapses that have fixed programmable weights and short-term plasticity dynamics (denoted as STP synapses). Since each of these synapse circuits can be stimulated individually by the chip input spikes, they can all operate in parallel using slow dynamics (e.g., driven by pico-Ampere currents), without having to transfer state memory at high speeds to external memory banks. Here, the area used by the parallel circuits allows to save bandwidth and power in implementing neural dynamics. Furthermore, the feature of implementing biologically plausible time constants and making use of explicit natural time scales, allows to use the fast digital asynchronous Address-Event Representation (AER) [7] circuits for stimulating multiple synapses in parallel.

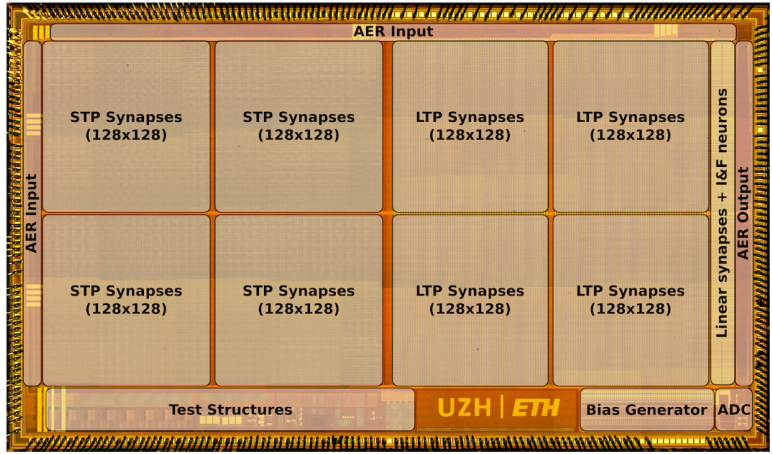

Fig. 3: Chip micrograph of the recurrent on-line learning (ROLLS) neuromorphic processor. Silicon neuron integrate and fire (I&F) circuits are placed on the right of the die area and are connected to an array of 256×512 synapses. Half of the synapse array is composed of learning synapses that exhibit Long Term Plasticity (LTP), and half is composed of fixed-weight synapses that exhibit Short Term Plasticity (STP). Input and output spikes are routed to the synapses and from the neurons via asynchronous digital Address-Event Representation (AER) circuits. On-chip DACs (bias generators) can be programmed to set the bias values of the analog synapse and neuron circuits. On-chip Analog to Digital Converters are used to read-out the currents from the current-mode DPI circuits that emulate synaptic dynamics.

In our work, we used the ROLLS device of Fig. 3 to enable learning in WTA networks to implement a hardware model of sequence learning [46], as well as learning maps [47] and sensorimotor mappings [48] in neuromorphic agents. Specifically, we used interconnected populations of spiking neurons in a Winner-Take-All (WTA) fashion, creating excitatory recurrent connections between neurons that encode similar features. When a group of such interconnected neurons is activated, the excitatory connections create an activation flow between neurons that can sustain itself even after the initiating input yields. Such sustained activation (or attractor) can drive down-stream structures, resulting in a movement of the agent; the time structure of this movement can be decoupled from the dynamics and timing of sensory input. Moreover, learning between sustained attractor states can be triggered when a rewarding or error signal is perceived.

The rather dense WTA connectivity requires parallel processing to be efficiently computed. Moreover, non-linearities of the DNF dynamics – which are crucial for its attractor properties – need to be processed for a large number of neurons in parallel. The explicit representation of space in neuromorphic devices, such as the one of Fig. 3, leads to a more efficient implementation of the neuronal attractor network, compared to hardware with time-multiplexing and virtual time. We thus argue that neuromorphic devices that feature real-time processing with time-scales matching the task and with massively parallel network of analog computing elements, when matched with a neuronal architecture of attractor-based population dynamics can lead to efficient implementations of neuromorphic systems capable to generate cognitive behavior. In the following Sections we present representative examples of such neuromorphic agents.

## III. Neuromorphic cognitive agents

In this section, we describe neuronal architectures that make use of the properties of neuromorphic devices with explicit representation of neuronal substrate and real time, using concepts of attractor dynamics in neuronal populations. We describe in detail two seminal architectures – a realisation of a closed-loop navigation controller inspired by a Braitenberg vehicle and a sequence learning architecture that makes use of plastic on-chip synapses to store sequences of observations, and review a few more examples where on-chip plasticity was used to learn along with behavior. Both these architectures were realized on the ROLLS neuromorphic processor. Despite of the small size of the network implemented, the neuromorphic processor is capable of produce robust behavior on a robotic agent in real-time.

### A. Closed-loop sensorimotor control

The first neuromorphic architecture demonstrates how behavior of a simple roving agent can be controlled by a spiking neuronal network. Such behavior can be generated by a very simple neuronal system, as has been demonstrated by V. Braitenberg with his classical "Vehicles" that generate meaningful behavior in a structured environment based on simple wiring of their sensors and motors [49]. According to this view, the most basic ability required to generate goal-directed behavior is the ability to differentiate between different sensory inputs. In simple terms, the agent needs to be able to tell one sensory state from another one. In terms of neuronal architecture, this means that the system needs to represent combinations of spatial and non-spatial features, in order to select the spatial target for a movement. Note that such combinations are also generated in deep convolutional neural network architectures in different feature maps. The spatial component, however, typically gets "lost" in over layers of the network that is trained for a recognition task. The first part of our architecture achieves such differentiation or representation of sensory states (note, we are showing a small scale example here on a chip with 256 spiking neurons).

Fig. 4 shows a neuronal architecture that realizes one of the Braitenberg's controllers on the neuromorphic device ROLLS. We used a WTA network to represent visual input obtained by a neuromorphic event-based camera Dynamic Vision Sensor (DVS), mounted on a miniature robotic vehicle Pushbot. In particular, neurons in two WTAs receive an external spike for each event in their receptive field: the "Target" and the "Obstacle" populations in the figure. Each physical neuron on the ROLLS that is assigned to belong to the Target WTA network is stimulated by visual events from the upper half of the DVS: Each neuron in the Target WTA "observes" a vertical column of the DVS frame. Because of the soft-WTA connectivity, activity bumps emerge in the target WTA that correspond to locations of salient visual inputs in the field of view (FoV) of the robot. In this simple example, the target direction is set by a blinking LED of the second robot, which can be observed in the upper half of the FoV. Objects in the lower half of the FoV are considered obstacles and drive the "Obstacles" neuronal population shown in the upper part of the architecture in Fig. 4. Thus, in the visual part of the controller, we differentiate between obstacles and targets based on the location of the input on the vertical axis of the FoV and we differentiate between locations (or directions) of input based on the horizontal coordinate of the DVS events.

Note that the physical instatiation of neurons on the ROLLS chip makes this architecture simple and elegant: no time-multiplexing and state-machine logic is needed, in fact no software-based arbitration or "algorithm" is used for processing. Instead, we have created a closed-loop dynamical system that processes sensory inputs and creates a representation in real, physical time. By connecting events from a camera to different neurons through synaptic weights that realize receptive fields of these neurons, we *represent* the visual input in neuronal substrate.

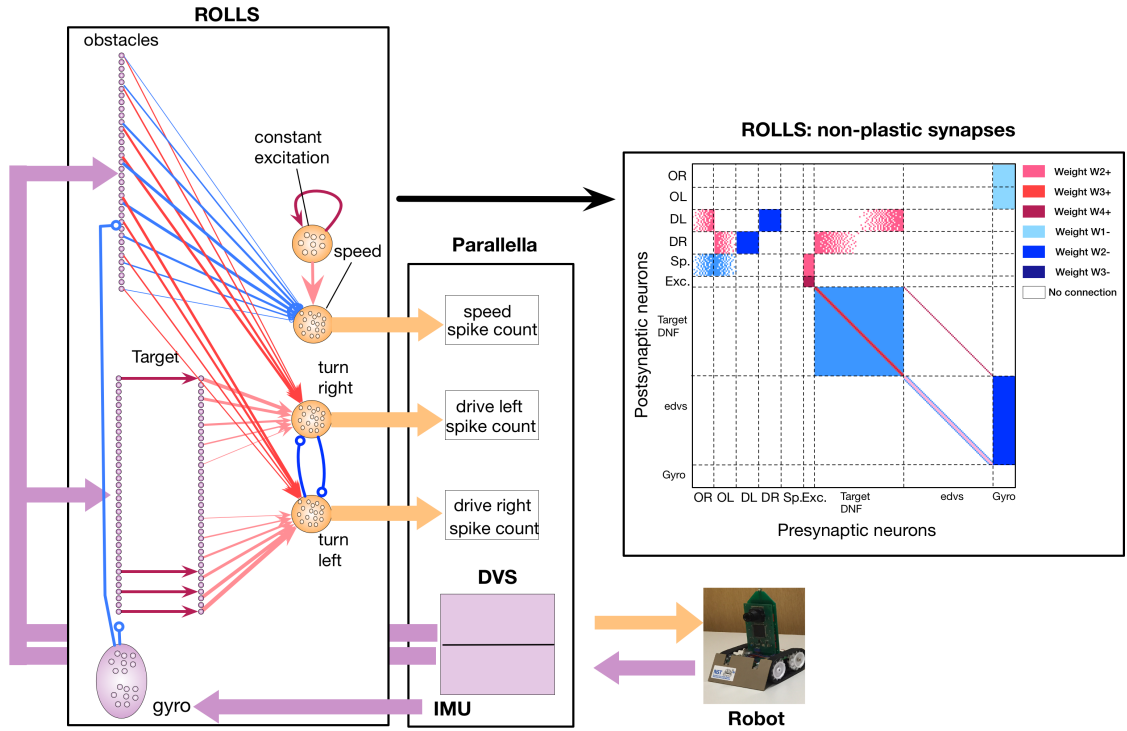

Fig. 4: **Left**: Scheme of the architecture for obstacle avoidance and target acquisition on the ROLLS chip. The architecture is realized on the ROLLS neuromorphic device with 256 adaptive mixed signal analog-digital leaky I&F neurons. The interface to the robot is established using a miniature computer board parallella. Output of the robot's visual sensor DVS is collected on the parallella and the DVS events are streamed to neurons on the ROLLS that represent horizontal coordinate of the events caused by either a target (upper half of the FoV) or an obstacle (lower half of the FoV). The second Target neuronal population has a strong WTA connectivity and selects a single target location. The motor populations (orange neuron groups) control the robot, setting its turning speed (to the right or left from its heading direction) and translational speed, respectively. The latter speed population is inhibited if an obstacles is detected in the Obstacle population, whereas the turning speeds are set by the excitatory connections from the Obstacles and Targets populations. Spikes from the motor neurons are collected on the parallella board and set the motor speeds of the robot. The "gyro" population inhibits visual neurons when the robot is turning, to compensate for additional events caused by turning ("saccadic suppression"). For more details see [50]. **Right**: The same architecture as connectivity matrix of non-plastic synapses on the ROLLS chip.

In this representation, each active neuron signals that there was some input coming from the portion of the sensor that this neuron observes. If several objects are present in front of a camera, several neurons will be active. In the Target population, moreover, the WTA connectivity makes sure that a single target location is selected and distractors are suppressed [50–52].

On the motor side, the visually induced representation of the location of potential obstacles represented by a neuronal population on the ROLLS chip is "wired" to neurons that represent movement. These movement neurons, if active, cause the robot to turn either to the right or to the left. Thus, each of the two motor populations in Fig. 4 activates a motor primitive that causes the robot's movement. In particular, the rotational velocity of the robot is set proportional to the firing rate of the respective (left or right) motor population. Presence of an obstacle also makes the robot to slow down, activity in the Obstacles population inhibits the "Speed" neuronal population. The firing rate of this latter population sets the robot's forward speed.

Overall, the simple neuronal architecture shown in Fig. 4 allows the robot to drive around, avoiding obstacles and approaching targets [50–52]. This behavior is produced by a dynamical system, created by the sensorimotor "embodiment" of the robot, situated in an environment, and the neuronal system that closes the behavior loop, connecting perception to motor control. This controller does not include any algorithmic arbitration: the sensory signals (DVS events and Gyroscope measurements) directly drive neurons and, eventually, the motors. The neuronal architecture ensures that a correct behavior is produced. The fact the neurons on the ROLLS chip are instatiated physically makes such efficient direct connection possible, reducing demands on the memory and arbitration that are usually managed by the CPU.

The neuronal architecture presented here does not include any learning. Indeed, this architecture is wired for the particular task, although the precise neuronal parameters can be found by a machine learning procedure [53], instead of human-labour tuning. In the next section, we present a neuronal architecture on the ROLLS chip that includes learning using plastic on-chip synapses. This architecture is one of our recent examples that show how representations – of temporal or spatial behaviorally-relevant patterns, such as sequences or maps, – can be learned using principles of attractor dynamics in hardware with explicit space and time.

*B. Learning sequences in a neuromorphic device*

Recently, we have implemented the basic serial order sequence learning model based on Dynamic Neural Fields [54] on the ROLLS chip [46]. In this architecture, a WTA population represents the content of the items in a sequence (here, location of a target on the horizontal axis of the DVS's FoV). Other neuronal populations represent serial order positions along the sequence: first, second, third, etc. ordinal group. Each group of these ordinal neurons is connected to the content WTA with plastic synapses that are updated when an ordinal neuron and a WTA neuron are active at the same time, according to a local synaptic plasticity rule. Groups of memory neurons create sustained activation and keep track of the position in the sequence in periods of sequential transitions, when ordinal groups are inhibited. This inhibition is induced by a condition-of-satisfaction (CoS) system – a group of neurons that is activated when an item in the sequence has been successfully executed [54, 55].

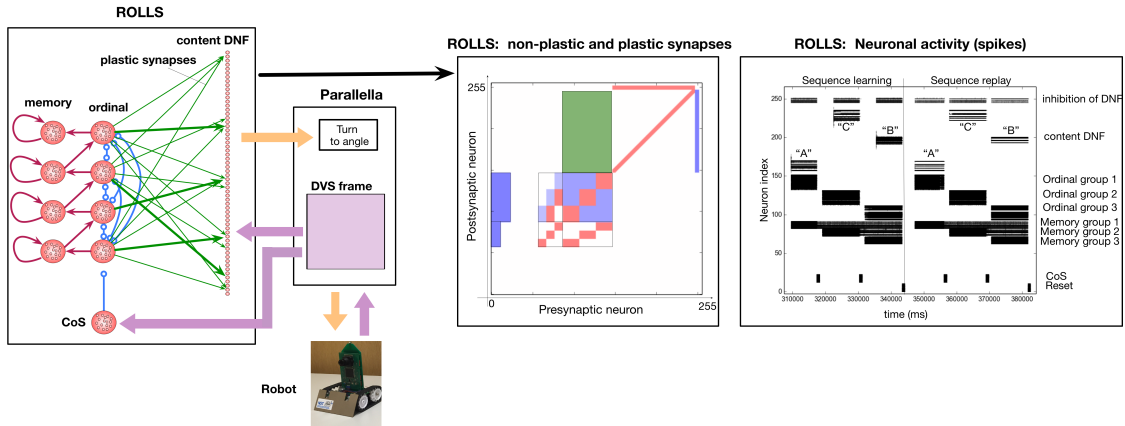

Fig. 5: Scheme of the architecture for sequence learning on the ROLLS chip and its realization with plastic and non-plastic synapses. Right plot shows a raster plot of neuronal activity in a sequence learning experiment, in which the robot observes three inputs, stores their order in plastic on-chip synapses and replays the items. See [47] and main text for details.

In [47] we have shown how sequence of states can be learned and replayed by a robotic agent, whose sensors and motors are interfaced to the neuromorphic device implementing the sequence learning architecture in Fig. 5. In this example, again, the physical identity of neurons is important to be able to efficiently, directly distribute activity patterns in the neuronal architecture. Moreover, the ability of the attractor networks to bridge different amounts of time is critical here: note that during sequence learning and production, each item in a sequence can take different - and not known in advance - time intervals. Neuronal dynamics can sustain these long time intervals, because stable activity bumps are created in the WTA neuronal population and in the ordinal groups. No clock is needed to keep track of time in these neuronal systems explicitly, although storing typical durations of actions is also possible in the neuronal population dynamics [56].

Another example of a neuro-dynamic architecture built using attractor networks and profiting from explicit representation of space and time in mixed signal neuromorphic devices is a neuromorphic self-localisation and map formation architecture [47, 57, 58]. In this system, first, a WTA neuronal population keeps track of the correct heading of the agent, performing path integration based on the motor commands sent to the robot. Another neuronal population keeps track of the position of the robot in an environment. The plastic synapses of the ROLLS chip are used to learn a simple map of the environment as the robot navigates it and senses walls with a bumper. The map is effectively an association between the representation of position (and orientation) of the robot and a neuronal representation of a collision. Learning happens in the moment when the event induced by the collision with an object activates the "Collision" population of neurons. Co-activation of these neurons and neurons that represent the current position in real time leads to increase of plastic synapses between these two groups, forming a collisions map as the robot explores the environment.

The same principle has been used to develop a neuromorphic PI controller that triggers learning of a mapping from the target speed of a robot to a motor command that realizes this speed [48]. Such mapping is an instantiation of a simple inverse model, which was learned using plastic synapses on the ROLLS chip. Here, learning was triggered when the PI controller (realized using the same principles of population dynamics and attractor WTA networks) converges and activates a zero-error neuronal population. Again, in this architecture sensory inputs drive neuronal representations in real time and learning is activated autonomously.

In all these neuromorphic architectures, a neuronal system is designed that connects sensors and motors to solve a particular task. Learning is reserved to certain portions of these architectures – e.g., a mapping between a sensory and motor space, or between representation of the ordinal position and perceptual content. The WTA connectivity of neuronal populations that represent perceptual states ensures that representations are localised and stabilised, limiting learning both in time and in space. In [57], we show how on-chip plasticity combined with WTA structure can also be used to learn to differentiate patterns in an unsupervised manner, while [48, 52] shows how neuronal populations can be used to perform arithmetic operations, such as summation and substraction, required, e.g., to realise a PI controller in a neural network or to perform reference frames transformations. The latter work emphasises that some computation that can easily be solved on a conventional von Neunmann computing platform might require rethinking and design of a neuronal architecture in neuromorphic computing devices.

## IV. Conclusion

Neuromorphic computing represents a promising paradigm for artificial cognitive systems that may become an enabling element for the design of power-efficient, real-time, and compact solutions for electronic devices that process the sensory data and generate behavior in real-world – changing and complex – environments. To minimize power consumption it is important to match the dynamics of neuromorphic computing elements to the time scale of physical processes that drive them and which the devices control. The impact of this approach is significant, because in addition to autonomous robots, "cognitive agents" represent also low-power always-on embedded sensory-computing systems (e.g., in cell-phones or wearable devices), intelligent wireless sensors (e.g., for Internet of Things -IoT- applications), or intelligent micro-bio-sensors (e.g., for personalized medicine, brain-machine interfaces, or prosthetics), or other types of micro-scale autonomous devices that extract structure from the data they acquire through their interactions with the environment and take decisions on how and when to act (to transmit information, power-up further processing stages, etc.)

Although many of the organizing principles of nervous systems have successfully been adopted for the design of artificial intelligence systems, two of the most powerful ones, i.e., that of letting time represent itself, and that of using physical instantiations of parallel circuits rather than single time-multiplexed ones, are not yet fully exploited. Here we presented examples of systems that implement also these principles and described a computational approach that extends the temporal processing abilities of such device to arbitrary time scales using attractor dynamics and the DNF framework. We have discussed several examples of neuro-dynamic architectures that make use of the physical instantiation of spiking neurons in hardware and their real-time dynamics, matched to biological time-scales. We believe these examples represent the first steps toward the realization of neuromorphic systems and computing architectures optimally suited for tasks in which behavior needs to be generated by autonomous neuromorphic agents in real-time, while staying adaptive in complex environments.

## Acknowledgments

Many of the concepts presented here were inspired by the discussions held at the CapoCaccia Cognitive Neuromorphic Engineering Workshop. The quantitative comparison of Table I was performed by Mohammad Ali Sharif Shazileh. This project has received funding from the European Research Council under the Grant Agreement No. 724295 (NeuroAgents) and the SNSF project Ambizione (PZ00P2_168183_1).